\documentclass{article} % For LaTeX2e
\usepackage{iclr2021_conference,times}

\usepackage{amsmath,amsfonts,bm}

\def\eqref#1{equation~\ref{#1}}
\def\1{\bm{1}}

\DeclareMathAlphabet{\mathsfit}{\encodingdefault}{\sfdefault}{m}{sl}
\SetMathAlphabet{\mathsfit}{bold}{\encodingdefault}{\sfdefault}{bx}{n}

\usepackage{hyperref}
\usepackage{url}
\usepackage{graphicx}
\usepackage{float}
\usepackage{chngcntr}
\usepackage[title]{appendix}

\iclrfinalcopy

\title{Goal-Directed Planning by Reinforcement Learning and Active Inference}

\author{Dongqi Han  \\
Cognitive Neurorobotics Research Unit\\
Okinawa Institute of Science and Technology\\
Okinawa, Japan \\
\And
Kenji Doya \\
Neural Computation Unit \\
Okinawa Institute of Science and Technology\\
Okinawa, Japan \\
\AND
Jun Tani\thanks{Correspondence: \texttt{jun.tani@oist.jp}} \\
Cognitive Neurorobotics Research Unit\\
Okinawa Institute of Science and Technology\\
Okinawa, Japan\\
}

\begin{document}

\maketitle

\begin{abstract}
What is the difference between goal-directed and habitual behavior? We propose a novel computational framework of decision making with Bayesian inference, in which everything is integrated as an entire neural network model. The model learns to predict environmental state transitions by self-exploration and generating motor actions by sampling stochastic internal states $\bm{z}$. Habitual behavior, which is obtained from the prior distribution of $\bm{z}$, is acquired by reinforcement learning. Goal-directed behavior is determined from the posterior distribution of $\bm{z}$ by planning, using active inference \textcolor{black}{which optimizes the past, current and future $\bm{z}$ by minimizing the variational free energy for the desired future observation constrained by the observed sensory sequence}. We demonstrate the effectiveness of the proposed framework by experiments in a sensorimotor navigation task with camera observations and continuous motor actions.
\end{abstract}

\section{Introduction}
\label{chap:introduction}
The mechanism of intelligent decision making is a central, frequently discussed problem in cognitive science, neuroscience, and artificial intelligence. \textcolor{black}{In particular, goal-directed planning, i.e., how to adaptively generate plans to achieve a non-fixed goal, has been of great interest since long ago \citep{duncan1996intelligence, tani1996model, desmurget1998eye}.}

The free energy principle (FEP) \citep{friston2010free} and active inference (AIf) theory \citep{friston2010action, friston2011action} provide a Bayesian computational framework of the brain. AIf explains decision-making as changing the agent's belief of perception and proprioception to minimize the difference (or \textit{surprise} in FEP) between predicted observation and goal observation \textcolor{black}{\citep{tani1996model, friston2016active, matsumoto2020goal}}. However, existing AIf studies are usually restricted to relatively simple environments with low-dimensional observation space and/or a known state transition model \citep{friston2009reinforcement,friston2017active, ueltzhoffer2018deep, millidge2020deep}. One major challenge is to the tradeoff between exploration and exploitation by simply minimizing the free energy \citep{ueltzhoffer2018deep, millidge2020deep, fountas2020deep}. 

On the other hand, deep reinforcement learning (RL) has been developed in recent years as a powerful framework for learning reward-oriented decision making. Goal-directed tasks can be converted to RL problems by providing a reward when the goal is achieved. Even though, modern deep RL has surpassed many single-objective tasks \citep{silver2017mastering, vinyals2019grandmaster, li2020suphx}, it is still challenging with multiple and/or non-fixed goals (reward function) \citep{plappert2018multi}. % Also, performance of RL depends highly on algorithms, hyperparameters, and reward design, which are usually empirically decided.

In this paper, we suggest that RL and AIf may be both necessary to achieve efficient and flexible goal-directed planning. We here propose a novel computational framework of goal-directed planning with Bayesian (variational) inference, in which RL and AIf work in a complementary manner. It performs deep RL for exploring the environment and acquiring habitual behavior for exploitation. Meanwhile, it learns a predictive model for sensory observation. Then, goal-directed planning can be performed under the framework of AIf by simply minimizing the free energy with respect to a given goal observation.

We demonstrate the effectiveness of the proposed framework in a simulated navigation task with high-dimensional observation space and continuous action space. We show that a flexible habitual behavior was acquired via RL for reward-seeking and collision avoidance, and that near-optimal goal-directed planning was obtained simply by optimizing the AIf objective.

\section{Methods}
\begin{figure}[t]
    \centering
    \includegraphics[width=0.75\textwidth]{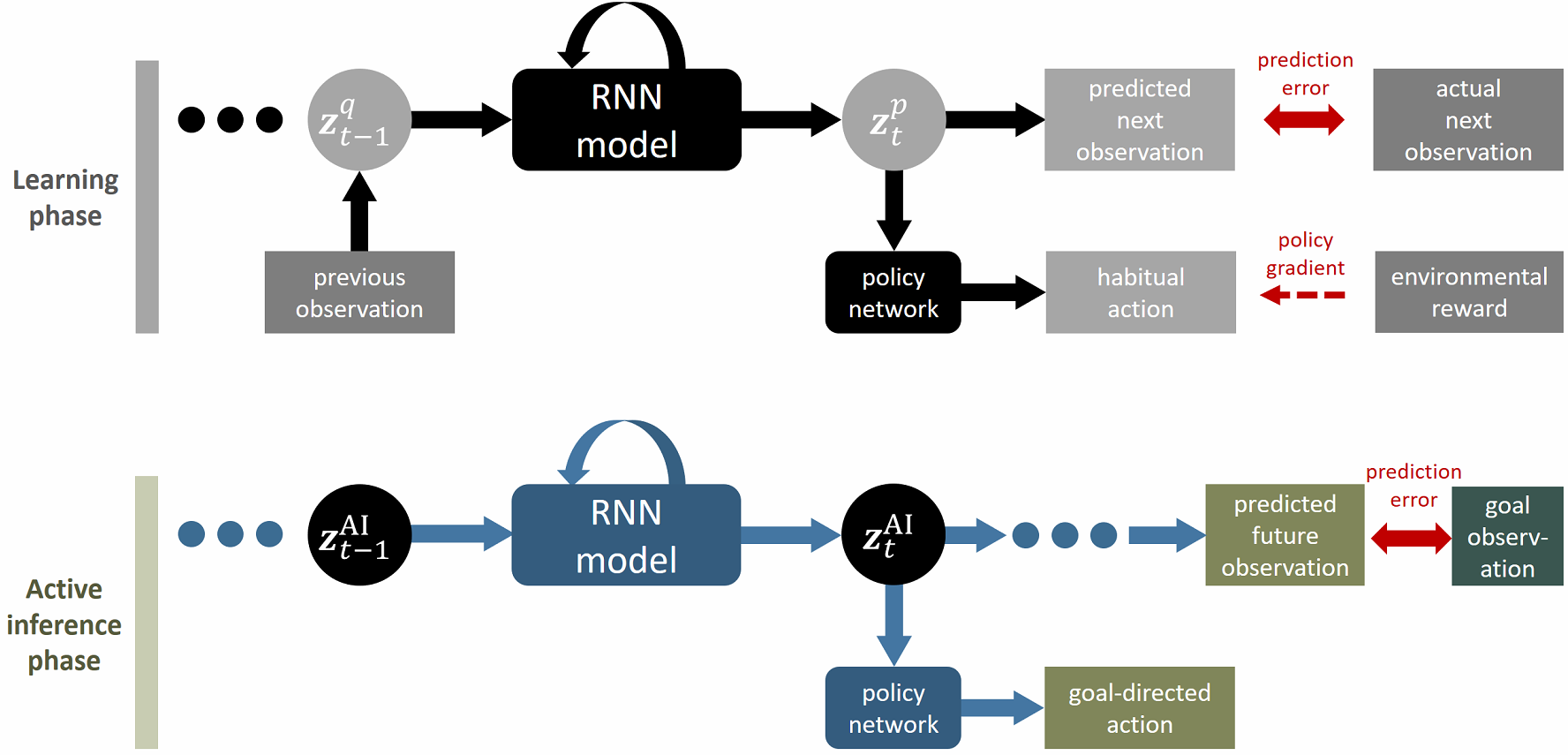}
    \caption{\textbf{Simplified diagrams of the the proposed framework.} In the learning phase, the prediction model and RL networks are trained. In the active inference phase, the internal states (latent variable) $\bm{z}^{\mbox{\tiny{AIf}}}$ are computed by error backpropagation to minimize the free energy for a given goal observation (Eq.~\ref{eq:er_loss}). Trainable variables in each phase are shown in black.}
    \label{fig:overview}
\end{figure}
\subsection{Overview}
\label{chap:overview}
The proposed framework employs an integrated neural network model that can perform RL and observation prediction simultaneously (Fig.~\ref{fig:overview}). The core idea is based on a variational Bayesian variable $\bm{z}$, referred to as \textit{internal states} in FEP \citep{friston2010free}. The  internal states $\bm{z}$ entail the probabilistic distribution of the agent's belief about action and perception. 

By exploring the environment, network connections are trained to predict observation transitions and to fulfill RL objectives, as shown in \textit{learning phase} of Fig.~\ref{fig:overview}. At time step $t$, the prior $\bm{z}^p_t$ is sampled by the RNN model and the motor action is generated via the policy network. The model also uses actual observations to infer the posterior $\bm{z}^q_{t-1}$ so that the RNN states reflect real environmental state (postdiction update). 

If the model is well learned, the agent can perform goal-directed planning using AIf, i.e., updating its internal states $z^{AIf}_t$ to minimize the error between the predicted future observation and the goal observation (Fig.~\ref{fig:overview}, \textit{AIf phase}). The motor action can then be obtained using the policy network as in the RL case, without requiring an inverse model.

The following sections detail computation processes of the proposed framework.

\begin{figure}
    \centering
    \includegraphics[width=1.0\textwidth]{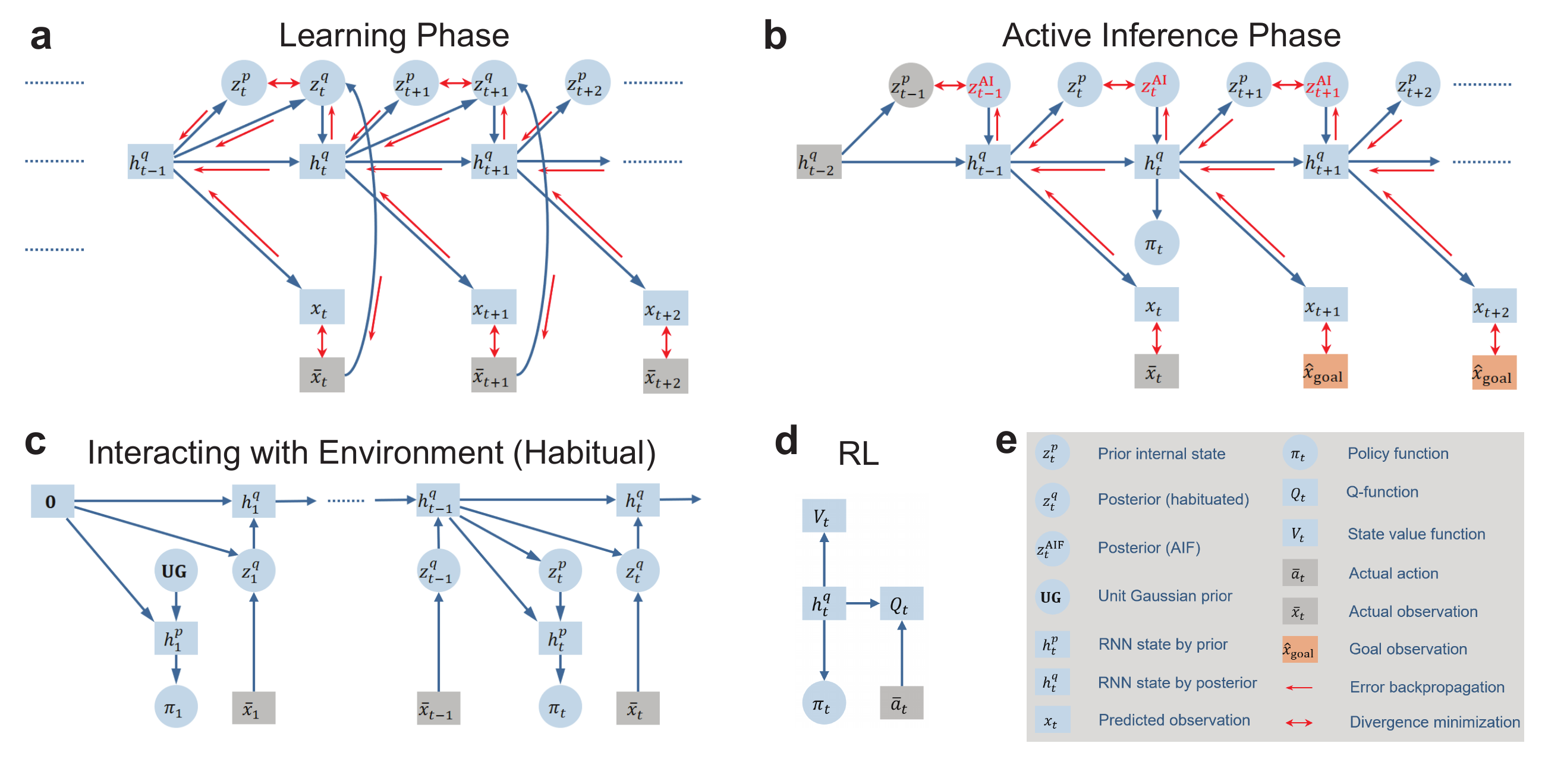}
    \caption{\textbf{Detailed diagrams of how the framework functions.} (a) Learning phase. (b) AIf phase, when current time step is $t$. (c) Interacting with the environment. (d) Computing RL functions. (e) Explanations of diagram nodes and edges. In (b),(c) and (d). Action $\bm{a}_t$ is sampled from the policy function $\pi_t$.}
    \label{fig:diagram}
\end{figure}

\subsection{Observation prediction model}
\label{chap:st_model}
To estimate a world model in general cases that are probably partially observable \citep{aastrom1965optimal}, \textcolor{black}{there exist various kinds of RNN-based models \citep{tani1996model, ha2018recurrent, kaiser2020model, han2020variational}.  Here we employ a Bayesian RNN model, as a natural choice under the FEP framework,} for predicting state transitions (Fig.~\ref{fig:diagram}(a)). It is an extension of the variational RNN (VRNN) model in \citet{chung2015recurrent}, and is explained as follows.

First, let $\bm{h}_{t}$ denote the RNN states, which are recurrently updated by
\begin{equation}
    \bm{h}_t = f^{\mbox{RNN}} (\bm{h}_{t-1}; \bm{z}^q_t),
\end{equation}
where we used the long short-term memory recurrency \citep{hochreiter1997long}, and $\bm{z^q}_t$ are posterior internal states which can be inferred by $\bm{h}_{t-1}$ and current raw observation $\bm{x}_t$:
\begin{equation}
    \bm{z}^q_t \sim \mathcal{N} \left( \bm{\mu}_{q,t}, \mbox{diag}(\bm{\sigma}^2_{q,t})  \right),  \quad \left[ \bm{\mu}_{q,t}, \bm{\sigma}^2_{q,t}  \right] = f^{\mbox{posterior}}(\bm{x}_t, \bm{h}_{t-1}).
    \label{eq:vrnn-inference}
\end{equation}
On the other hand, prior internal states $\bm{z}^p_t$ can be obtained from $\bm{h}_{t-1}$ solely:
\begin{equation}
    \bm{z}^p_t \sim \mathcal{N} \left( \bm{\mu}_{p,t}, \mbox{diag}(\bm{\sigma}^2_{p,t})  \right),\quad  \left[ \bm{\mu}_{p,t}, \bm{\sigma}^2_{p,t}  \right] = f^{\mbox{prior}}(\bm{h}_{t-1}) .
\end{equation}
The model predicts the next observation (for pixel observation) as
\begin{equation}
 \bm{x}_{t+1}  = \mbox{sigmoid}\left( \bm{\mu}_{x,t+1}  \right),\quad  \bm{\mu}_{x,t+1} = f^{\mbox{decoder}}(\bm{h}_t).
\end{equation}
The free energy (or negative variational lower bound) objective can be written as 
\begin{equation}
    F = \sum_t^T\left[D_{KL} (q(\bm{z}_t) || p(\bm{z}_t)\right] - \mathbb{E}_{q} \left[ \log \left( p(\bm{x}_t = \bar{\bm{x}}_t)\right) \right],  \label{eq:model_loss}
\end{equation}
where $p$ and $q$ are parameterized PDFs of prior and posterior $\bm{z}_t$, respectively, using the reparameterization trick \citep{kingma2013auto}. We estimate $\log \left( p(\bm{x}_t = \bar{\bm{x}}_t)\right) $ by cross entropy, assuming a Bernoulli distribution of $\bm{x}_t$. The model is trained by minimizing Eq.~\ref{eq:er_loss} by backpropagation through time.  See Appendix~\ref{append:imple_details} for more details.

\subsection{Reinforcement learning}
\label{chap:rl}
Fig.~\ref{fig:diagram}(c) demonstrates how action is obtained in habitual behavior. It is worth mentioning that the connection from $\bm{z}^p_t$ to $\bm{h}^p_t$ shares the same synaptic weight with the connection from $\bm{z}^q_t$ to $\bm{h}^q_t$. This is consistent with the fact that $\bm{z}^p_t, \bm{h}^p_t$ and $\bm{z}^q_t, \bm{h}^q_t$ represent two sides of the same coin, but the posterior has additional information about the actual observation $\bar{\bm{x}}_t$.

We used soft actor-critic (SAC) as the RL algorithm \citep{haarnoja2018soft, haarnoja2019soft}. As shown in Fig.~\ref{fig:diagram}(d), the state value function, policy function and state-action value function are estimated by $f_V(\bm{h})$, $f_\pi(\bm{h})$ and $f_Q(\bm{h}, \bm{a})$, respectively, where $f_V$, $f_\pi$ and $f_Q$ are fully-connected feedforward networks. Note that during learning, gradients backpropagate (through time) to the whole RNN model. This is beneficial to shape representations of RNN states that are useful for motor control (more details can be found in Appendix~\ref{append:rl}).

\subsection{Active inference by error regression}
\label{chap:aif}
After learning a predictive model and control skills by RL, the proposed framework is able to perform goal-directed planning. As shown in  (Fig.~\ref{fig:diagram}(b)), there is a goal observation provided and AIf is conducted by minimizing the corresponding free energy by updating internal states $\bm{z}^{\mbox{\tiny{AIf}}}$. More specifically, we do \textit{error regression}\citep{ahmadi2017bridging, ahmadi2019novel, matsumoto2020goal}, which is to minimize the error between predicted observation and goal observation regularized by KL divergence between prior and posterior, that is, the free energy with respect to the goal observation: 
\begin{align}
    F^{\mbox{\tiny{AIf}}} = &  D_{KL} (q^{\mbox{\tiny{AIf}}}(\bm{z}_{t-1}) || p(\bm{z}_{t-1})) - \log \left(p(\bm{x}_{t}=\bar{\bm{x}}_t) \right)+ \nonumber\\ & \sum_{\tau=0}^{N-1}\left[ D_{KL} (q^{\mbox{\tiny{AIf}}}(\bm{z}_{t+\tau}) ||  p(\bm{z}_{t+\tau})) - c_\tau \log \left( p(\bm{x}_{t+\tau+1}=\bm{x}_{goal})\right) \right].  \label{eq:er_loss}
\end{align}
The posterior internal states by AIf $\bm{z}^{\mbox{\tiny{AIf}}}_{t}$ is computed using gradient descent and backpropagation through time to minimize Eq.~\ref{eq:er_loss} at each time step $t$ (see Appendix~\ref{append:er}). The previous-step posterior $\bm{z}^{\mbox{\tiny{AIf}}}_{t-1}$ is also trained so as to represent the current true environmental state (the first line of Eq.~\ref{eq:er_loss}).

Note that the prediction loss at future steps is multiplied by trainable scalar variables $c_\tau$ where $\tau=0,1,\cdots,N-1$ (we used $N=8$). This is because the number of steps by which the agent can reach the goal is unknown; therefore, we introduce $c_\tau$, which can be understood as the probability of reaching the goal at $t + \tau$. Softmax is used so that we have $c_0 + c_1 + \cdots + c_{N-1} = 1$.

\section{Results}
\begin{figure}
    \centering
    \includegraphics[width=1.0\textwidth]{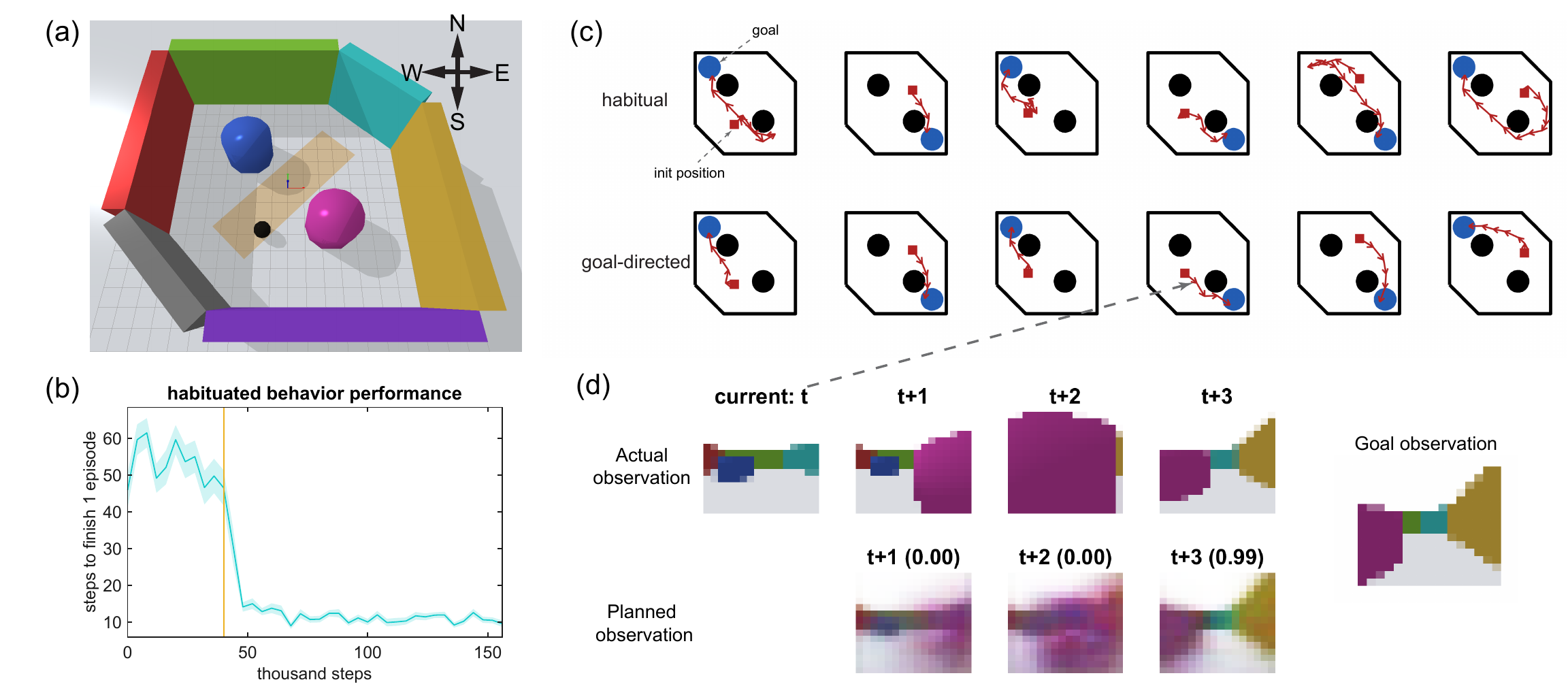}
    \caption{\textbf{Results of the maze navigation task.} (a) The environment in the PyBullet simulator, where the black ball is the robot and other objects are walls and obstacles. In each episode, the robot is initialized at a random position in the orange, shaded area in the middle. (b) The learning curve of habitual behavior (Mean $\pm$ MSE., 40 trials). The vertical line indicates the start of learning, before which random exploration is conducted. (c) Comparison of moving trajectories between habitual and goal-directed behavior. (d) Actual observation and predicted observation in AIf, in an example episode (the northern camera is shown). Numbers in the parentheses indicate $c_\tau$ in Eq.~\ref{eq:er_loss}, i.e. the possibility of achieving the goal at that the corresponding time step.}
    \label{fig:results}
\end{figure}
We tested the proposed framework in a simulated maze environment (Fig.~\ref{fig:results}(a)). In this task, the goal is to reach the reward area at the southeast or northwest corner while bypass the obstacles, at which the agent can receive a one-step reward and finish the episode. There is also a punishment (negative reward) given if the robot collides with any wall or obstacle.

The robot makes observations using RGBD camera images. For simplicity, we mounted 4 cameras on the robot, each of which has a fixed orientation (east, south, west, north) so the observation depends only on the robot location. The robot can move on a plane by executing a continuous-value motor action $(\Delta X, \Delta Y)$ with a limited maximum distance at each step.

In each episode, the reward area is randomly set at either the southeast or northwest corner, and the robot cannot see the reward area. Therefore, if no clue about the goal is available to the agent, it should try to reach both reward areas. This is habituated behavior. In contrast, if a goal position to be reached is specified as the sensory observation expected at the goal position, the agent should directly go to the corresponding position, which is goal-directed behavior. Using the proposed framework, we performed simulations on this task.

First, the agent explored the environment with stochastic actions and collected data (observation, action and reward sequences) in its replay buffer. The model was trained by minimizing Eq.~\ref{eq:model_loss} and RL objectives with experience replay. It acquired habitual behavior by first going to one goal area and then to the other, if the task is not completed (Fig.~\ref{fig:results}(b) and the upper row of (c)).

If a goal is assigned and the observation at the goal position is known to the agent, goal-directed planning can be conducted using the method described in Sect.~\ref{chap:aif}. To demonstrate goal-directed behavior, we visualized the moving trajectories of goal-directed behavior of an example agent (after learning) in the lower row of Fig.~\ref{fig:results}(c), compared to trajectories of habitual behavior in the upper row. The agent could plan actions directly toward the desired goal, in contrast to the frequent redundant moving in habitual behavior, due to lack of goal clue. We can also see that the goal-directed behavior shared good motor-control skills of habitual behavior, such as avoiding collisions or zigzagging. We also demonstrate how the agent predicted future observations that lead to the given goal in Fig.~\ref{fig:results}(d).

\section{Summary}

This work is a proof of concept illustrating how RL and AIf collaborate in a single, integrated network model to perform habitual behavior learning and goal-directed planning with high-dimensional observation space and continuous action space. Our key idea is to utilize the variational internal states $\bm{z}$. The prior distribution $\bm{z}^p$ corresponds to unconditional, habitual behavior for exploration/exploitation. Goal-directed behavior is obtained by computing the posterior $\bm{z}^{\mbox{\tiny{AIf}}}$ \textcolor{black}{(including deciding current $\bm{z}^{\mbox{\tiny{AIf}}}_t$ and refreshing previous $\bm{z}^{\mbox{\tiny{AIf}}}_{t-1}$) using AIf to minimize the expected free energy} for a given goal observation. The effectiveness of our framework is demonstrated using a maze navigation task with pixel images as observations and continuous motor actions.

Future work will scale up the model to more realistic and challenging environments. Also, it is interesting to consider how habitual and goal-directed behavior switch without artificial assignment, which may underlie an important mechanism of decision-making in the brain.

\section*{Acknowledgement}
The authors are funded by OIST graduate school and this material is based on work that is partially funded by an unrestricted gift from Google.

\bibliography{iclr2021_conference}

\begin{thebibliography}{32}
\providecommand{\natexlab}[1]{#1}
\providecommand{\url}[1]{\texttt{#1}}
\expandafter\ifx\csname urlstyle\endcsname\relax
  \providecommand{\doi}[1]{doi: #1}\else
  \providecommand{\doi}{doi: \begingroup \urlstyle{rm}\Url}\fi

\bibitem[Ahmadi \& Tani(2017)Ahmadi and Tani]{ahmadi2017bridging}
Ahmadreza Ahmadi and Jun Tani.
\newblock Bridging the gap between probabilistic and deterministic models: a
  simulation study on a variational {B}ayes predictive coding recurrent neural
  network model.
\newblock In \emph{International Conference on Neural Information Processing},
  pp.\  760--769. Springer, 2017.

\bibitem[Ahmadi \& Tani(2019)Ahmadi and Tani]{ahmadi2019novel}
Ahmadreza Ahmadi and Jun Tani.
\newblock A novel predictive-coding-inspired variational rnn model for online
  prediction and recognition.
\newblock \emph{Neural computation}, pp.\  1--50, 2019.

\bibitem[{\AA}str{\"o}m(1965)]{aastrom1965optimal}
Karl~J {\AA}str{\"o}m.
\newblock Optimal control of {M}arkov processes with incomplete state
  information.
\newblock \emph{Journal of Mathematical Analysis and Applications}, 10\penalty0
  (1):\penalty0 174--205, 1965.

\bibitem[Chung et~al.(2015)Chung, Kastner, Dinh, Goel, Courville, and
  Bengio]{chung2015recurrent}
Junyoung Chung, Kyle Kastner, Laurent Dinh, Kratarth Goel, Aaron~C Courville,
  and Yoshua Bengio.
\newblock A recurrent latent variable model for sequential data.
\newblock In \emph{Advances in neural information processing systems}, pp.\
  2980--2988, 2015.

\bibitem[Coumans \& Bai(2016--2019)Coumans and Bai]{coumans2019pybullet}
Erwin Coumans and Yunfei Bai.
\newblock Pybullet, a python module for physics simulation for games, robotics
  and machine learning.
\newblock \url{http://pybullet.org}, 2016--2019.

\bibitem[Desmurget et~al.(1998)Desmurget, P{\'e}lisson, Rossetti, and
  Prablanc]{desmurget1998eye}
Michel Desmurget, Denis P{\'e}lisson, Yves Rossetti, and Claude Prablanc.
\newblock From eye to hand: planning goal-directed movements.
\newblock \emph{Neuroscience \& Biobehavioral Reviews}, 22\penalty0
  (6):\penalty0 761--788, 1998.

\bibitem[Duncan et~al.(1996)Duncan, Emslie, Williams, Johnson, and
  Freer]{duncan1996intelligence}
John Duncan, Hazel Emslie, Phyllis Williams, Roger Johnson, and Charles Freer.
\newblock Intelligence and the frontal lobe: The organization of goal-directed
  behavior.
\newblock \emph{Cognitive psychology}, 30\penalty0 (3):\penalty0 257--303,
  1996.

\bibitem[Fountas et~al.(2020)Fountas, Sajid, Mediano, and
  Friston]{fountas2020deep}
Zafeirios Fountas, Noor Sajid, Pedro~AM Mediano, and Karl Friston.
\newblock Deep active inference agents using {M}onte-{C}arlo methods.
\newblock In \emph{Advances in neural information processing systems}, 2020.

\bibitem[Friston(2010)]{friston2010free}
Karl Friston.
\newblock The free-energy principle: a unified brain theory?
\newblock \emph{Nature reviews neuroscience}, 11\penalty0 (2):\penalty0
  127--138, 2010.

\bibitem[Friston et~al.(2011)Friston, Mattout, and Kilner]{friston2011action}
Karl Friston, J{\'e}r{\'e}mie Mattout, and James Kilner.
\newblock Action understanding and active inference.
\newblock \emph{Biological cybernetics}, 104\penalty0 (1):\penalty0 137--160,
  2011.

\bibitem[Friston et~al.(2016)Friston, FitzGerald, Rigoli, Schwartenbeck,
  Pezzulo, et~al.]{friston2016active}
Karl Friston, Thomas FitzGerald, Francesco Rigoli, Philipp Schwartenbeck,
  Giovanni Pezzulo, et~al.
\newblock Active inference and learning.
\newblock \emph{Neuroscience \& Biobehavioral Reviews}, 68:\penalty0 862--879,
  2016.

\bibitem[Friston et~al.(2017)Friston, FitzGerald, Rigoli, Schwartenbeck, and
  Pezzulo]{friston2017active}
Karl Friston, Thomas FitzGerald, Francesco Rigoli, Philipp Schwartenbeck, and
  Giovanni Pezzulo.
\newblock Active inference: a process theory.
\newblock \emph{Neural computation}, 29\penalty0 (1):\penalty0 1--49, 2017.

\bibitem[Friston et~al.(2009)Friston, Daunizeau, and
  Kiebel]{friston2009reinforcement}
Karl~J Friston, Jean Daunizeau, and Stefan~J Kiebel.
\newblock Reinforcement learning or active inference?
\newblock \emph{PloS one}, 4\penalty0 (7):\penalty0 e6421, 2009.

\bibitem[Friston et~al.(2010)Friston, Daunizeau, Kilner, and
  Kiebel]{friston2010action}
Karl~J Friston, Jean Daunizeau, James Kilner, and Stefan~J Kiebel.
\newblock Action and behavior: a free-energy formulation.
\newblock \emph{Biological cybernetics}, 102\penalty0 (3):\penalty0 227--260,
  2010.

\bibitem[Ha \& Schmidhuber(2018)Ha and Schmidhuber]{ha2018recurrent}
David Ha and J\"{u}rgen Schmidhuber.
\newblock Recurrent world models facilitate policy evolution.
\newblock In S.~Bengio, H.~Wallach, H.~Larochelle, K.~Grauman, N.~Cesa-Bianchi,
  and R.~Garnett (eds.), \emph{Advances in Neural Information Processing
  Systems 31}, pp.\  2450--2462. Curran Associates, Inc., 2018.

\bibitem[Haarnoja et~al.(2018{\natexlab{a}})Haarnoja, Zhou, Abbeel, and
  Levine]{haarnoja2018soft}
Tuomas Haarnoja, Aurick Zhou, Pieter Abbeel, and Sergey Levine.
\newblock Soft actor-critic: Off-policy maximum entropy deep reinforcement
  learning with a stochastic actor.
\newblock In \emph{International Conference on Machine Learning}, pp.\
  1856--1865, 2018{\natexlab{a}}.

\bibitem[Haarnoja et~al.(2018{\natexlab{b}})Haarnoja, Zhou, Hartikainen,
  Tucker, Ha, Tan, Kumar, Zhu, Gupta, Abbeel, et~al.]{haarnoja2019soft}
Tuomas Haarnoja, Aurick Zhou, Kristian Hartikainen, George Tucker, Sehoon Ha,
  Jie Tan, Vikash Kumar, Henry Zhu, Abhishek Gupta, Pieter Abbeel, et~al.
\newblock Soft actor-critic algorithms and applications.
\newblock \emph{arXiv preprint arXiv:1812.05905}, 2018{\natexlab{b}}.

\bibitem[Han et~al.(2020{\natexlab{a}})Han, Doya, and Tani]{han2020self}
Dongqi Han, Kenji Doya, and Jun Tani.
\newblock Self-organization of action hierarchy and compositionality by
  reinforcement learning with recurrent neural networks.
\newblock \emph{Neural Networks}, 129:\penalty0 149--162, 2020{\natexlab{a}}.

\bibitem[Han et~al.(2020{\natexlab{b}})Han, Doya, and Tani]{han2020variational}
Dongqi Han, Kenji Doya, and Jun Tani.
\newblock Variational recurrent models for solving partially observable control
  tasks.
\newblock In \emph{Proceedings of the International Conference on Learning
  Representations}, 2020{\natexlab{b}}.

\bibitem[Hochreiter \& Schmidhuber(1997)Hochreiter and
  Schmidhuber]{hochreiter1997long}
Sepp Hochreiter and Jurgen Schmidhuber.
\newblock Long short-term memory.
\newblock \emph{Neural computation}, 9\penalty0 (8):\penalty0 1735--1780, 1997.

\bibitem[Kaiser et~al.(2020)Kaiser, Babaeizadeh, Milos, Osinski, Campbell,
  Czechowski, Erhan, Finn, Kozakowski, Levine, et~al.]{kaiser2020model}
Lukasz Kaiser, Mohammad Babaeizadeh, Piotr Milos, Blazej Osinski, Roy~H
  Campbell, Konrad Czechowski, Dumitru Erhan, Chelsea Finn, Piotr Kozakowski,
  Sergey Levine, et~al.
\newblock Model-based reinforcement learning for atari.
\newblock In \emph{Proceedings of the International Conference on Learning
  Representations (ICLR)}, 2020.

\bibitem[Kingma \& Ba(2014)Kingma and Ba]{kingma2014adam}
Diederik~P Kingma and Jimmy Ba.
\newblock Adam: A method for stochastic optimization.
\newblock \emph{arXiv preprint arXiv:1412.6980}, 2014.

\bibitem[Kingma \& Welling(2013)Kingma and Welling]{kingma2013auto}
Diederik~P Kingma and Max Welling.
\newblock Auto-encoding variational {B}ayes.
\newblock \emph{arXiv preprint arXiv:1312.6114}, 2013.

\bibitem[Li et~al.(2020)Li, Koyamada, Ye, Liu, Wang, Yang, Zhao, Qin, Liu, and
  Hon]{li2020suphx}
Junjie Li, Sotetsu Koyamada, Qiwei Ye, Guoqing Liu, Chao Wang, Ruihan Yang,
  Li~Zhao, Tao Qin, Tie-Yan Liu, and Hsiao-Wuen Hon.
\newblock Suphx: Mastering mahjong with deep reinforcement learning.
\newblock \emph{arXiv preprint arXiv:2003.13590}, 2020.

\bibitem[Matsumoto \& Tani(2020)Matsumoto and Tani]{matsumoto2020goal}
Takazumi Matsumoto and Jun Tani.
\newblock Goal-directed planning for habituated agents by active inference
  using a variational recurrent neural network.
\newblock \emph{Entropy}, 22\penalty0 (5):\penalty0 564, 2020.

\bibitem[Millidge(2020)]{millidge2020deep}
Beren Millidge.
\newblock Deep active inference as variational policy gradients.
\newblock \emph{Journal of Mathematical Psychology}, 96:\penalty0 102348, 2020.

\bibitem[Plappert et~al.(2018)Plappert, Andrychowicz, Ray, McGrew, Baker,
  Powell, Schneider, Tobin, Chociej, Welinder, et~al.]{plappert2018multi}
Matthias Plappert, Marcin Andrychowicz, Alex Ray, Bob McGrew, Bowen Baker,
  Glenn Powell, Jonas Schneider, Josh Tobin, Maciek Chociej, Peter Welinder,
  et~al.
\newblock Multi-goal reinforcement learning: Challenging robotics environments
  and request for research.
\newblock \emph{arXiv preprint arXiv:1802.09464}, 2018.

\bibitem[Silver et~al.(2017)Silver, Schrittwieser, Simonyan, Antonoglou, Huang,
  Guez, Hubert, Baker, Lai, Bolton, et~al.]{silver2017mastering}
David Silver, Julian Schrittwieser, Karen Simonyan, Ioannis Antonoglou, Aja
  Huang, Arthur Guez, Thomas Hubert, Lucas Baker, Matthew Lai, Adrian Bolton,
  et~al.
\newblock Mastering the game of go without human knowledge.
\newblock \emph{Nature}, 550\penalty0 (7676):\penalty0 354, 2017.

\bibitem[Tani(1996)]{tani1996model}
Jun Tani.
\newblock Model-based learning for mobile robot navigation from the dynamical
  systems perspective.
\newblock \emph{IEEE Transactions on Systems, Man, and Cybernetics, Part B
  (Cybernetics)}, 26\penalty0 (3):\penalty0 421--436, 1996.
\newblock \doi{10.1109/3477.499793}.

\bibitem[Ueltzh{\"o}ffer(2018)]{ueltzhoffer2018deep}
Kai Ueltzh{\"o}ffer.
\newblock Deep active inference.
\newblock \emph{Biological cybernetics}, 112\penalty0 (6):\penalty0 547--573,
  2018.

\bibitem[Uhlenbeck \& Ornstein(1930)Uhlenbeck and
  Ornstein]{uhlenbeck1930theory}
George~E Uhlenbeck and Leonard~S Ornstein.
\newblock On the theory of the brownian motion.
\newblock \emph{Physical review}, 36\penalty0 (5):\penalty0 823, 1930.

\bibitem[Vinyals et~al.(2019)Vinyals, Babuschkin, Czarnecki, Mathieu, Dudzik,
  Chung, Choi, Powell, Ewalds, Georgiev, et~al.]{vinyals2019grandmaster}
Oriol Vinyals, Igor Babuschkin, Wojciech~M Czarnecki, Micha{\"e}l Mathieu,
  Andrew Dudzik, Junyoung Chung, David~H Choi, Richard Powell, Timo Ewalds,
  Petko Georgiev, et~al.
\newblock Grandmaster level in starcraft ii using multi-agent reinforcement
  learning.
\newblock \emph{Nature}, 575\penalty0 (7782):\penalty0 350--354, 2019.

\end{thebibliography}
\bibliographystyle{iclr2021_conference}

% \appendix
\begin{appendices}
\counterwithin{figure}{section}

%--------------------------------
% Below is appendix
%---------------------------------

\section{}

\subsection{Task details}
\label{append:mazee}
\begin{figure}[h]
    \centering
    \includegraphics[width=0.6\textwidth]{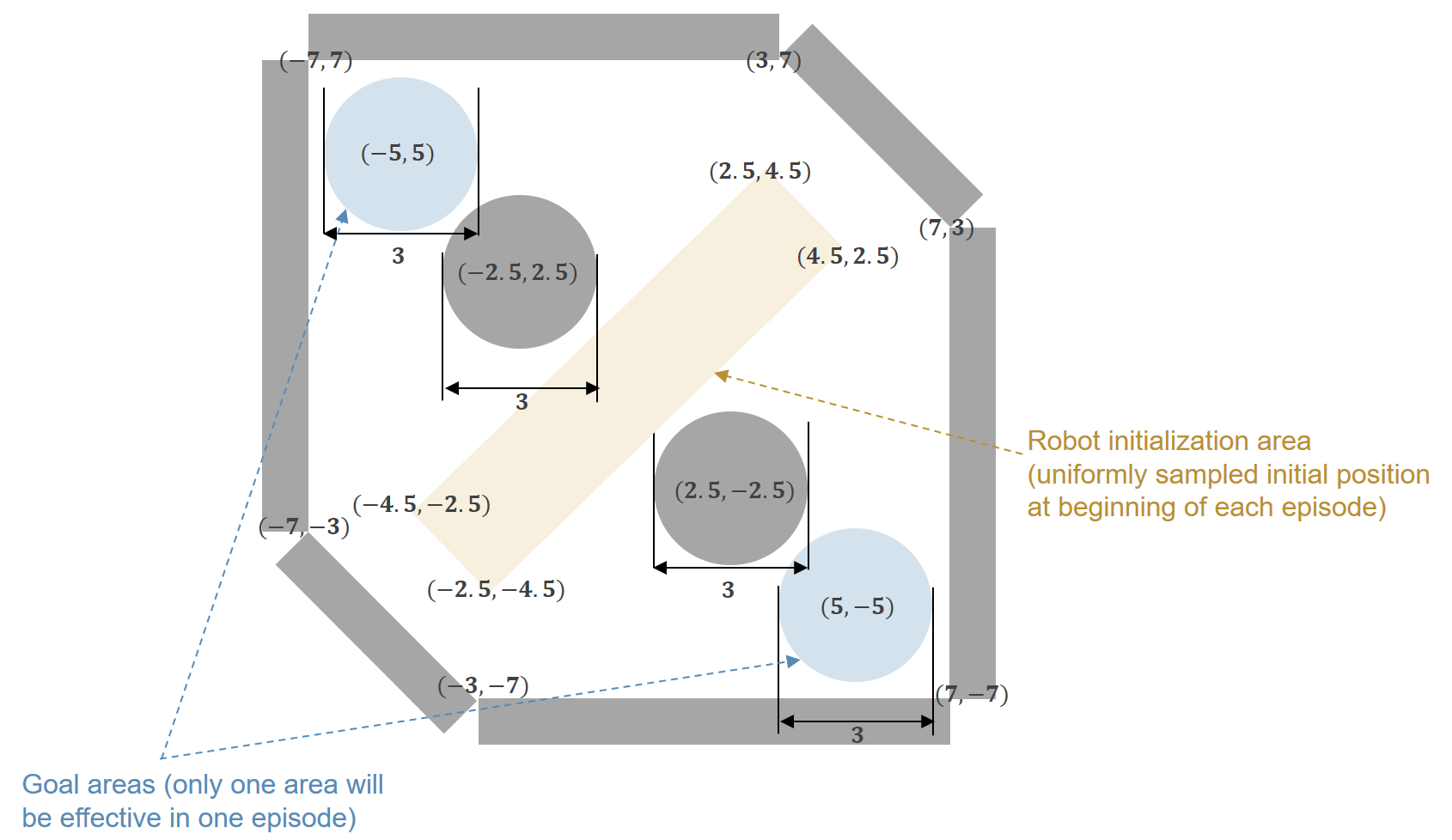}
    \caption{Configuration of the robot navigation task, where the numbers indicate ($x$,$y$) coordinates of the corresponding points.}
    \label{fig:mazee}
\end{figure}
We use PyBullet physics simulator for the experiments \citep{coumans2019pybullet}. Detailed configuration of the task we used can be seen in Fig.~\ref{fig:mazee}. The robot is a sphere with radius=0.5. And its action is its speed on the horizontal plane: $\Delta x$ and $\Delta y$, where $\Delta x \in [-2, 2]$ and $\Delta y \in [-2, 2]$ are bounded continuous variables. 

Observation of the robot is 4 RGBD cameras, each with resolution 16$\times$16. All the cameras are fixed on the top of the robot, and their directions are toward east, south, west and north, respectively (all perpendicular to the $z$ axis). The depth channel is normalized so that the depth value is bounded in [0, 1] as RGB channels. We treat the entire observation as a 12-channels 16$\times$16 image.

At each episode, one random goal area (out of two) will be effective. Once the center of the robot reaches the goal area, a reward of 100 is given. If the robot collides with any wall or obstacle, it receives a negative reward of -2.5.

\subsection{Implementation details}
\label{append:imple_details}

\subsubsection{Model architecture}
\label{append:model}
\begin{figure}
    \centering
    \includegraphics[width=1.0\textwidth]{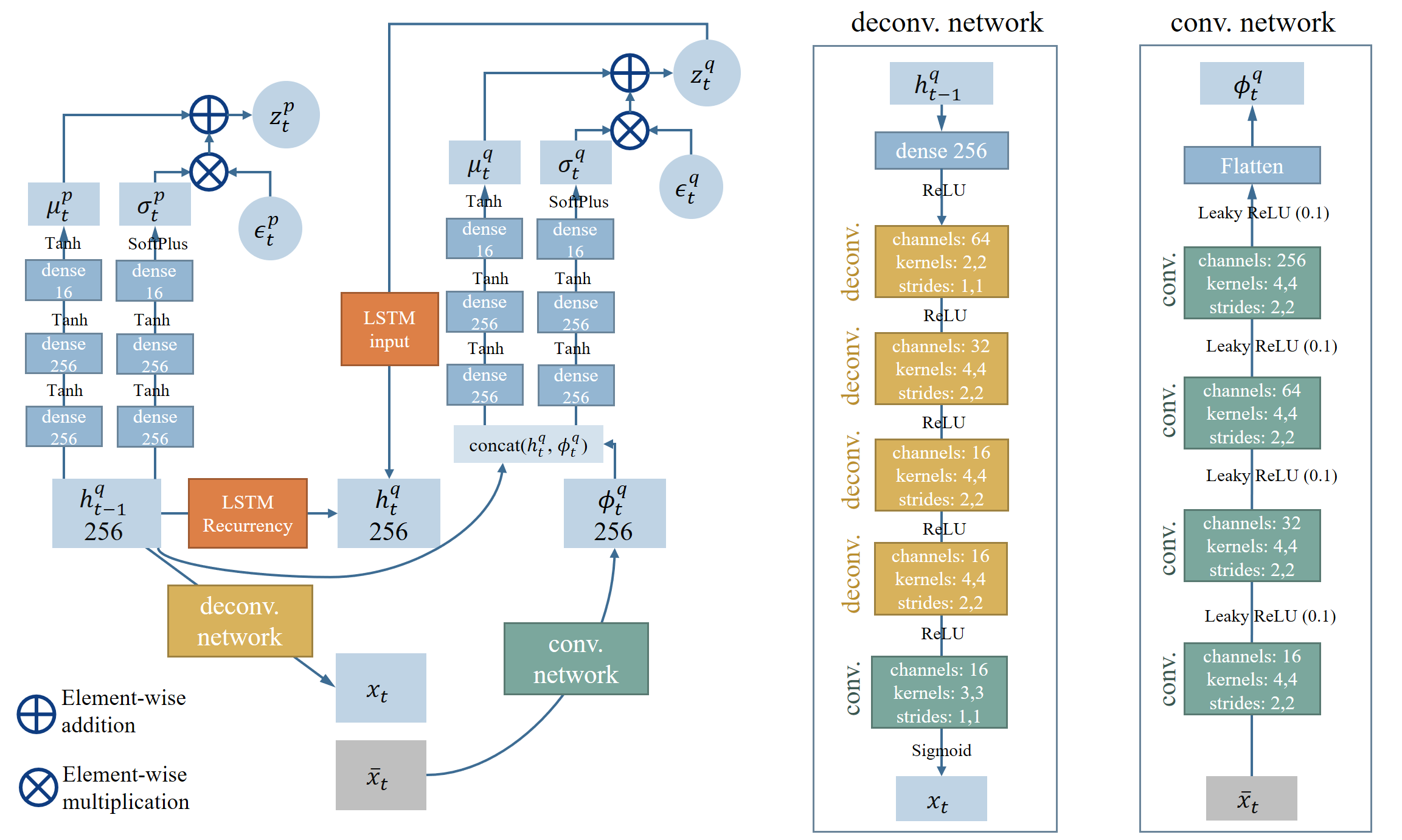}
    \caption{Architecture of our variational RNN model, where $\epsilon_t^p$ and $\epsilon_t^q$ are unit Gaussian white noise vector, which has the same dimension as the internal states $\bm{z}_t$.}
    \label{fig:model_archi}
\end{figure}

The detailed architecture of the VRNN model (not including RL networks) is shown in Fig.~\ref{fig:model_archi}. The RL networks are the same as in \citet{haarnoja2019soft} except that the input is not raw observation, but RNN states $\bm{h}_t$. Detailed diagrams of the model in different phases are shown in Fig.~\ref{fig:diagram}.

\subsubsection{Model hyperparameters}
To train the model for predicting next observation, Eq.~\ref{eq:model_loss} was minimized using observation transition data from the replay buffer. The replay buffer recorded observation transition data ($\bm{x}_t$) of recent 2,500 episodes (it also records $\bm{a}_t, r_t, done_t$ for the sake of RL).  The data used for each training step is 8 length-16 sequences of $\bm{x}_t$, randomly sampled from the replay buffer. We used learning rate 0.0003 and Adam optimizer \citep{kingma2014adam}. A training step was performed every 3 time steps.

\subsubsection{RL algorithm}
\label{append:rl}
We used discount factor $\gamma=0.8$ for RL. Reply buffer and batch size settings were the same as in model training. A gradient step was performed every 3 time steps. Note that for better exploration, we used Ornstein–Uhlenbeck process \citep{uhlenbeck1930theory} for generating motor noise, where the inverse-timescale constant $\theta_{\mbox{\tiny{OU}}}=0.3$, similar to that in \citet{han2020self}.

Other parameters followed the original SAC (with adaptive entropy regularization) implementation \citet{haarnoja2019soft}.

\subsubsection{Error regression}
\label{append:er}
In active inference phase, the goal observation was given by the environment, which is the expected observation at the center of a randomly selected goal area (it can also be sampled from the agent's replay buffer). When computing the posterior $\bm{z}^{\mbox{\tiny{AIf}}}_t$ that minimizes the free energy for the goal observation \citep{ahmadi2019novel, matsumoto2020goal}, all the model weights were fixed. At each time step, trainable variables $\bm{A}^\mu_\tau$ and $\bm{A}^\sigma_\tau$ (mean and standard deviation of the Gaussian distribution of $\bm{z}^{\mbox{\tiny{AIf}}}_{t+\tau}$) were randomly initialized, where $\tau=-1,0,1,\cdots,7$ (which means searching for 1 previous step and 8 future steps, see Eq.~\ref{eq:er_loss} and Fig.~\ref{fig:diagram}(b)). There were also trainable variables $\bm{A}^c_\tau$ initialized as zeros, where $\tau=0,1,\cdots,7$ and $c_\tau=\mbox{softmax}_{\tau=0,\cdots,7}(\bm{A}^c_\tau)$ (see Sect.~\ref{chap:aif}). Then, we had $\bm{z}^{\mbox{\tiny{AIf}}}_{t+\tau} =  \bm{\mu}^{\mbox{\tiny{AIf}}}_\tau = \mbox{tanh}(\bm{A}^\mu_\tau)$ and $ \bm{\sigma}^{\mbox{\tiny{AIf}}}_\tau = \mbox{softplus}(\bm{A}^\sigma_\tau)$ so that the free energy Eq.~\ref{eq:er_loss} could be computed.

We used batch size=16 (which means a batch of plans with different random initialization) so that there was a higher chance that at least one in the batch converging to the desired plan. We trained the batch of $\bm{A}^\mu_\tau$, $\bm{A}^\sigma_\tau$ and $\bm{A}^c_\tau$ using a RMSProp optimizer with decay=0.9 and learning rate=0.005, for 1,000 training steps at every actual time step. To choose a better plan from the batch, we first filtered out the half with higher free energies for goal observation. Then we estimated the expected time steps to reach the goal of each planned path by $l=\sum_{\tau=0}^7 c_\tau(\tau+1)$. Finally, we selected the plan in the batch with smallest $l$ (smallest expected time steps to reach the goal) to obtain $\bm{z}^{\mbox{\tiny{AIf}}}_{t} = \mbox{tanh}(\bm{A}^\mu_0)$ and the goal-directed policy could be obtained from $\bm{z}^{\mbox{\tiny{AIf}}}_{t}$ via the policy network. 

\end{appendices}
\end{document}